\documentclass{article}

\usepackage[final]{nips_2017}

\usepackage[utf8]{inputenc} 
\usepackage[T1]{fontenc}    
\usepackage{hyperref}       
\usepackage{url}            
\usepackage{booktabs}       
\usepackage{amsfonts}       
\usepackage{nicefrac}       
\usepackage{microtype}      

\usepackage{amsmath}
\usepackage{blkarray} 
\usepackage{dsfont} 
\usepackage{graphicx} 
\usepackage{bm} 
\usepackage[]{placeins} 

\usepackage[numbers]{natbib}
\makeatletter
\def\bbordermatrix#1{\begingroup \m@th
  \global\let\perhaps@scriptstyle\scriptstyle
  \@tempdima 4.75\p@
  \setbox\z@\vbox{%
    \def\cr{%
      \crcr
      \noalign{%
        \kern2\p@
        \global\let\cr\endline
        \global\let\perhaps@scriptstyle\relax
      }%
    }%
    \ialign{$\make@scriptstyle{##}$\hfil\kern2\p@\kern\@tempdima
      &\thinspace\hfil$\perhaps@scriptstyle##$\hfil
      &&\quad\hfil$\perhaps@scriptstyle##$\hfil\crcr
      \omit\strut\hfil\crcr
      \noalign{\kern-\baselineskip}%
      #1\crcr\omit\strut\cr}}%
  \setbox\tw@\vbox{\unvcopy\z@\global\setbox\@ne\lastbox}%
  \setbox\tw@\hbox{\unhbox\@ne\unskip\global\setbox\@ne\lastbox}%
  \setbox\tw@\hbox{$\kern\wd\@ne\kern-\@tempdima\left[\kern-\wd\@ne
    \global\setbox\@ne\vbox{\box\@ne\kern2\p@}%
    \vcenter{\kern-\ht\@ne\unvbox\z@\kern-\baselineskip}\,\right]$}%
  \null\;\vbox{\kern\ht\@ne\box\tw@}\endgroup}
\def\make@scriptstyle#1{\vcenter{\hbox{$\scriptstyle#1$}}}
\makeatother

\title{Modeling sepsis progression using hidden Markov models}

\author{
Brenden K. Petersen \\
Lawrence Livermore National Laboratory \\
Livermore, California 94550 \\
\texttt{petersen33@llnl.gov}
\And
Michael B. Mayhew \\
Lawrence Livermore National Laboratory \\
Livermore, California 94550 \\
\texttt{mmayhew@inflammatix.com} \\
\And
Kalvin O. E. Ogbuefi \\
Lawrence Livermore National Laboratory \\
Livermore, California 94550 \\
\texttt{kalvin.ogbuefi@student.csulb.edu} \\
 \And
John D. Greene \\
Kaiser Permanente Northern California \\
Oakland, California 94611 \\
\texttt{john.d.greene@kp.org} \\
\And
Vincent X. Liu \\
Kaiser Permanente Northern California \\
Oakland, California 94611 \\
\texttt{vincent.x.liu@kp.org} \\ 
 \And
Priyadip Ray \\
Lawrence Livermore National Laboratory \\
Livermore, California 94550 \\
\texttt{ray34@llnl.gov} \\
}

\begin{document}

\maketitle

\begin{abstract}
Characterizing a patient's progression through stages of sepsis is critical for enabling risk stratification and adaptive, personalized treatment. However, commonly used sepsis diagnostic criteria fail to account for significant underlying heterogeneity, both between patients as well as over time in a single patient. We introduce a hidden Markov model of sepsis progression that explicitly accounts for patient heterogeneity. Benchmarked against two sepsis diagnostic criteria, the model provides a useful tool to uncover a patient's latent sepsis trajectory and to identify high-risk patients in whom more aggressive therapy may be indicated.

 
\end{abstract}

\section{Introduction}

Sepsis is a dysregulated immune response to infection, accounting for nearly 50\% of all hospital deaths and \$15 billion in United States healthcare costs \citep{Liu2014}. Understanding a patient's progression through stages of sepsis is a prerequisite for effective risk stratification and adaptive, personalized treatment.


Hidden Markov models (HMMs) are a popular technique for modeling disease progression \citep{Jackson2003}. Early efforts on modeling sepsis progression based on Markov models \citep{Frausto1998, Brause2002} consider patients as a homogeneous group. However, septic patients' presentation of the condition and response to treatment are often characterized by significant heterogeneity both between patients and over time in a single patient \citep{Marshall2005}. Moreover, current clinical diagnostic criteria (e.g. sepsis-1 criteria \citep{Bone1992, Mayo} and quick sequential organ failure assessment (qSOFA) \citep{Singer2016}) only take into account levels (not variability) of a few vital signs---typically at a single time point---without considering the past status of the patient.

We introduce a discrete-time HMM to analyze patient progression through sepsis. Each discrete time step is associated with five vital signs: systolic blood pressure, diastolic blood pressure, heart rate, respiratory rate, and temperature. We account for patient heterogeneity to determine state transition probabilities, incorporating patient age and two composite measures of illness as covariates. We leverage a retrospective cohort of 25,000 patients with suspected or confirmed infection, a subset of the Kaiser Permanente Northern California dataset \citep{Liu2014}.



\section{Model specification}

\begin{figure}
\begin{center}
\includegraphics[trim={0 2cm 0 4.5cm}, clip]{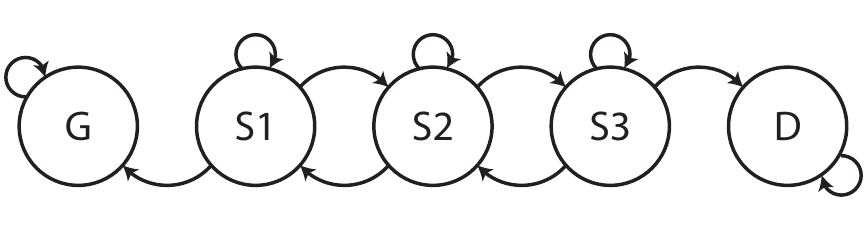}
\caption{Sepsis progression Markov process. Latent states $S1$, $S2$, and $S3$ represent increasing severity. Discharged ($G$) and death ($D$) are absorbing states.}
\label{fig:process}
\end{center}
\end{figure}

The hidden Markov process is summarized in Fig. \ref{fig:process}. It comprises five states: discharged ($G$), three latent states of increasing severity ($S1$, $S2$, and $S3$), and death ($D$). Both $G$ and $D$ are absorbing states and are observed. Transition probabilities are based on a proportional hazards model: for patient $i$, three covariates (age, acute physiology score (LAPS2), and chronic disease burden score (COPS2)), denoted $\bm{c}^{(i)}$, determine baseline risk. Global parameters $\bm{\beta}$, $\lambda_k$, and $\gamma_k$ $(k \in \{S1, S2, S3\})$ control the balance of improving from state $k$ (i.e. moving left in Fig. \ref{fig:process}), worsening from state $k$ (moving right), or remaining in state $k$. The transition probability matrix for patient $i$ is: 
\[
\bbordermatrix{
& \scriptstyle G & \scriptstyle S1 & \scriptstyle S2 & \scriptstyle S3 & \scriptstyle D \cr
\scriptstyle G	&	1 				&	0				&	0				&	0				&	0		\cr
\scriptstyle S1	&	\gamma_1P_1		&	(1-\gamma_1)P_1	&	1-P_1			&	0				&	0		\cr
\scriptstyle S2	&	0				&	\gamma_2P_2		&	(1-\gamma_2)P_2	&	1-P_2			&	0		\cr
\scriptstyle S3	&	0				&	0				&	\gamma_3P_3		&	(1-\gamma_3)P_3	&	1-P_3	\cr
\scriptstyle D	&	0				&	0				&	0				&	0				&	1		\cr
},
\]

\noindent where $P_k \equiv \lambda_k\exp(-\bm{\beta}^\top \bm{c}^{(i)})$.


Transient states ($S1$, $S2$, and $S3$) are associated with five patient vital signs, denoted $\bm{x}_t^{(i)}$ for patient $i$ at time interval $t$. Vital signs were pre-processed to occur at synchronous six-hour time intervals. Given a latent state $k \in \{S1, S2, S3\}$, the emission probability for a patient's five vital signs are modeled as normally distributed with mean vector $\bm{\mu}_k$ and a diagonal covariance matrix $\bm{\Sigma}_k$. Since states $G$ and $D$ are observed, they are not associated with vital signs. 


\section{Results and Discussion}

\subsection{Model inference}

We implemented a Metropolis-Hastings-within Gibbs sampler. Latent states ($z_t^{(i)}$), healing parameters ($\gamma_k$), and emission means ($\bm{\mu}_k$) and covariance matrices ($\bm{\Sigma}_k$) are inferred with Gibbs updates. (We omit derivations for brevity.) We placed a discrete uniform prior on $z_t^{(i)}$, a $\mathcal{U}(0, 1)$ prior on $\gamma_k$, a flat $\textrm{Inv-Gamma}(0.001, 1000)$ prior on diagonal elements in $\Sigma_k$, and normal priors on entries in $\bm{\mu}_k$ for which posterior distributions were insensitive to prior hyperparameter choices. The remaining parameters, $\bm{\beta}$ and $\lambda_k$, were inferred with Metropolis-Hastings updates. We placed an improper $\textrm{Gamma}(0, 0)$ prior for each element of $\bm{\beta}$ and a $\textrm{Beta}(100, 2)$ prior for $\lambda_k$.

We estimated all model parameters on a dataset of 20,000 patient hospitalization episodes. The sampler generated 10,000 samples, by which point the chains were well-mixed. The last 2,000 samples were used to construct posterior distributions. We estimated the marginal maximum a posteriori (MAP) value of each continuously valued parameter by applying a kernel density estimate to the posterior samples of that parameter and estimating its mode. MAP estimates for emission distribution means and standard deviations are shown in Table \ref{table:emissions}.

\begin{table}[h]
  \caption{Marginal maximum a posteriori estimates for mean and standard deviation of emission distributions for each vital sign and from each latent state.}
  \label{table:emissions}
  \centering
  \begin{tabular}{lccc}
    \toprule
    & \multicolumn{3}{c}{Latent state} \\
    \cmidrule{2-4}
    Vital sign & $S1$ & $S2$ & $S3$ \\
    \midrule
    
Systolic blood pressure (mm Hg) & $118.6 \pm 15.1$ & $143.4 \pm 16.3$ & $116.4 \pm 17.5$ \\
Diastolic blood pressure (mm Hg) & $63.4 \pm 9.3$ & $77.2 \pm 10.0$ & $62.7 \pm 11.2$ \\
Heart rate (min$^{-1}$) & $76.7 \pm 12.1$ & $83.3 \pm 14.5$ & $95.6 \pm 16.4$ \\
Respiratory rate (min$^{-1}$) & $18.7 \pm 1.6$ & $19.1 \pm 1.9$ & $21.1 \pm 4.9$ \\
Temperature ($^\circ$F) & $98.0 \pm 0.8$ & $98.1 \pm 0.8$ & $98.6 \pm 1.3$ \\
    
    \bottomrule
  \end{tabular}
\end{table}

\subsection{Model validation}

For model validation and subsequent analyses, we leveraged two commonly used clinical sepsis diagnostic criteria. Firstly, an early consensus definition developed sepsis criteria (later termed "sepsis-1") based primarily on patient vital signs that was used for several decades \citep{Bone1992, Mayo}. Sepsis-1 criteria are defined as bacterial infection plus two or more of the following systemic inflammatory response syndrome (SIRS) conditions: (1) heart rate > 90 min$^{-1}$, (2) respiratory rate > 20 min$^{-1}$ or PaCO$_2$ < 32 mm Hg, (3) temperature < 96.8 $^\circ$F or temperature > 100.4 $^\circ$F, and (4) white blood cell count > 12,000/mm$^3$ or < 4,000/mm$^3$ or >10\% immature bands. Note that the bacterial infection criteria is met for all patients in the dataset. Secondly, a more recent consensus sepsis definition involves a sequential organ failure assessment, or SOFA score, based primarily on laboratory test results \citep{Singer2016}. A simplified version of this assessment, called "quick SOFA" or qSOFA, includes two vital signs (systolic blood pressure $\leq$ 100 mm Hg, respiratory rate $\geq$ 22) that can be directly applied to our dataset at any time point. 

We applied the vital sign portions of sepsis-1 and qSOFA criteria to each time point for all patients, resulting in segments that indicate when diagnostic criteria are met. To validate the model, we assessed overlaps between the inferred $S3$ segments of each patient and segments indicated by the two diagnostic criteria. For both sepsis-1 and qSOFA criteria, the presence (or absence) of sepsis diagnosis tended to increase (or decrease) with increasing severity of the HMM states. We also note that inferred emission means from state $S3$ are consistent with vital sign values in sepsis-1 criteria.


\subsection{Mortality risk analysis}

The HMM's most severe state ($S3$) provides clinical utility in mortality risk analysis. To assess the ability of the $S3$ state to identify patients at high risk of mortality during hospitalization, we first inferred the latent state trajectories of a held-out 5,000-patient dataset. For this dataset, global parameters were fixed to the previously found MAP estimates, and the HMM was not provided with patient outcomes. To compare the ability of sepsis-1 criteria, qSOFA criteria, and the $S3$ state to distinguish between discharged and deceased patients, for each patient we calculated the proportion of time points for which sepsis-1 or qSOFA criteria were met, or for which the predicted state was $S3$. We then plotted these distributions, conditioned on patient outcome (Fig. \ref{fig:histograms}). The Jensen-Shannon divergence for each resulting distribution pair was 0.168 for sepsis-1 criteria, 0.097 for qSOFA criteria, and 0.186 for the $S3$ state, indicating that the inferred $S3$ state is more discriminative of the underlying risk state of the patient.



\subsection{Sepsis trajectory analysis}

\begin{figure}[h]
\begin{center}
\includegraphics[width=\textwidth,height=\textheight,keepaspectratio, trim={1.7cm 9.4cm 1.8cm 1.1cm}, clip]{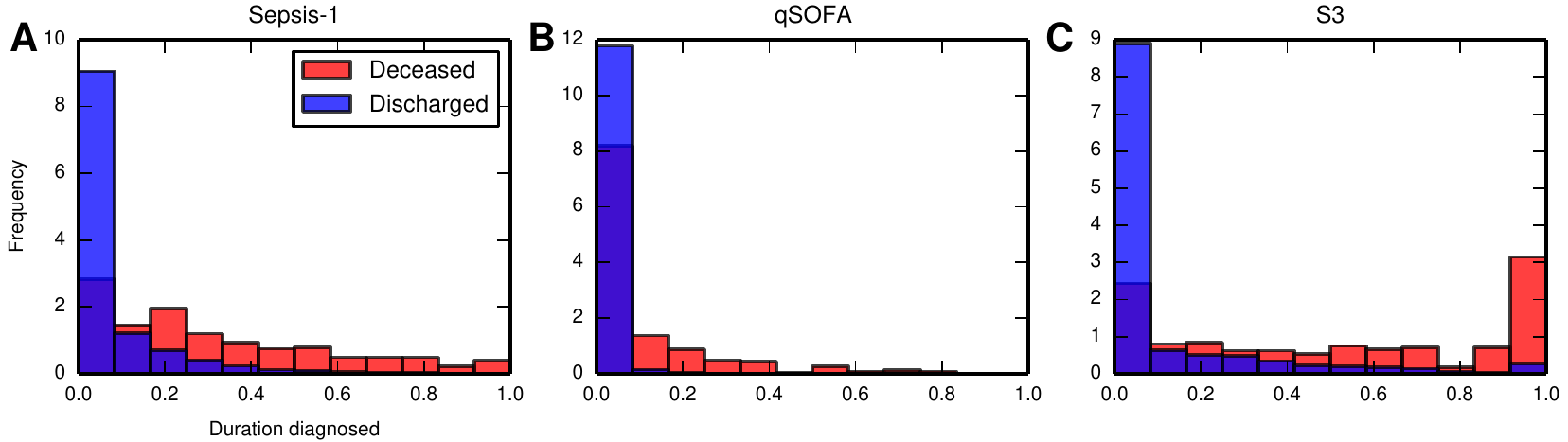}
\caption{Distributions over each patient's fraction of hospitalization duration for which sepsis-1 (A) or qSOFA (B) criteria were met, or for which the predicted state was $S3$ (C), conditioned upon patient outcome. Distributions are normalized to have unit area.}
\label{fig:histograms}
\end{center}
\medskip
\begin{center}
\includegraphics[width=\textwidth,height=\textheight,keepaspectratio, trim={1.5cm 0.1cm 1.7cm 1.5cm}, clip]{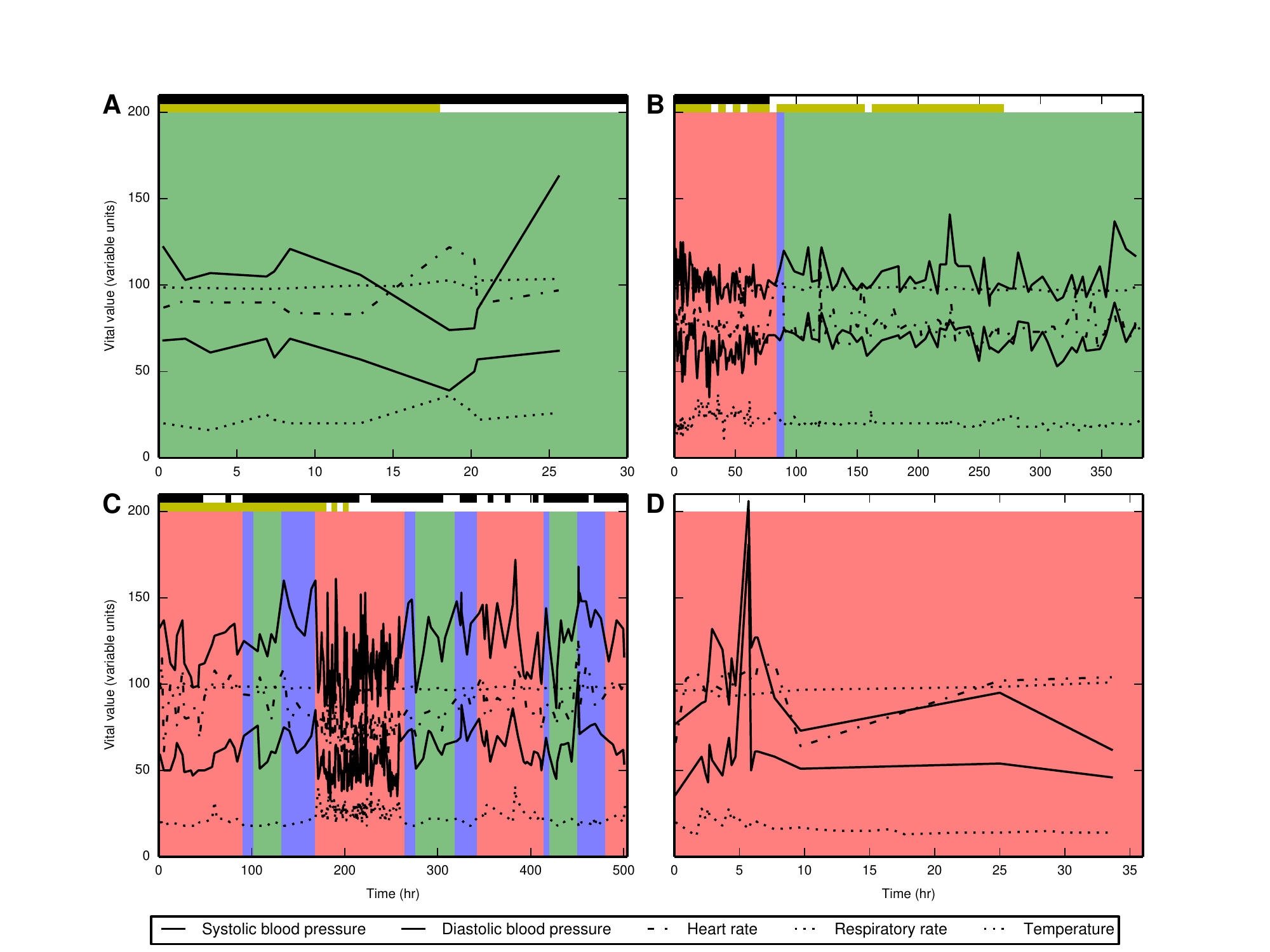}
\caption{Characteristic state trajectories for four patients. Time intervals are colored according to the marginal maximum a posteriori latent state (green = $S1$, blue = $S2$, red = $S3$). Five vital sign time series are overlaid: systolic blood pressure (mm Hg), diastolic blood pressure (mm Hg), heart rate (min$^{-1}$), respiratory rate (min$^{-1}$), and temperature ($^{\circ}$F). Two sets of bars (black or yellow) along the top of the trajectory indicate time segments for which sepsis-1 (black bar) or qSOFA (yellow bar) criteria diagnose the patient as septic. For A and B, the patient was discharged; for C and D, the patient died.}
\label{fig:states}
\end{center}
\end{figure}

Visualizing patient trajectories provides insights into underlying physiological patterns. We overlay a patient's vital signs with their predicted (MAP estimate) latent state for each time interval. Four characteristic trajectories are shown in Fig. \ref{fig:states}. In contrast to the HMM's $S3$ state, sepsis-1 and/or qSOFA criteria often over-predict sepsis for a patient who is ultimately discharged (Fig. \ref{fig:states}A), or even predict sepsis at time of discharge (Fig. \ref{fig:states}B). Conversely, both criteria can miss diagnoses for the entire trajectory of patients who die (Fig. \ref{fig:states}D). The Markovian nature of the HMM also provides a degree of temporal smoothing to the state trajectory, whereas sepsis-1 and qSOFA criteria can repeatedly alternate diagnoses within a short timeframe, as in Fig. \ref{fig:states}B-C.

\FloatBarrier

Thus, while sepsis-1 and qSOFA criteria may prove useful in an early phase of hospitalization as a trigger for clinical action, they have limited utility for representing the dynamic state of a patient's clinical course. Our HMM addresses these limitations by modeling temporal dependence and incorporating patient-specific features, allowing for adaptive, personalized characterization of patients' physiologic states within a trajectory toward recovery or deterioration.


\section{Conclusion}


This study demonstrates the utility in using HMMs to provide insight into a patient's underlying physiological trajectory, which may be used to inform clinical decisions. Including additional physiologic, laboratory, and treatment data in such models will likely improve the identification of high-risk patients in whom more aggressive therapy is indicated as well as low-risk patients in whom hospital discharge is indicated.

\FloatBarrier

\subsubsection*{Acknowledgments}

Dr. Vincent X. Liu is funded by an NIH grant (NIH K23GM112018). This work was performed under the auspices of the U.S. Department of Energy by Lawrence Livermore National Laboratory under Contract DE-AC52-07NA27344 (LLNL-CONF-740757).

\bibliography{sample}

\end{document}